\documentclass{article}
\usepackage{booktabs}
\usepackage{spconf,amsmath,amssymb,graphicx,hyperref}

\title{SAMP: Self-Adaptive Margin Prioritization for Robust Diffusion Watermarking}

\name{Yuhao Xue$^{\star1}$ \qquad Jiuan Zhou$^{\star1,2}$ \qquad Yu Cheng$^{1,2}$ \qquad Zhaoxia Yin$^{\dagger1}$}

\address{$^{1}$East China Normal University\\
$^{2}$Shanghai Innovation Institute\\
$^{\star}$Equal contribution\\
$^{\dagger}$Corresponding author: zxzyin@cee.ecnu.edu.cn}

\begin{document}
\ninept
\maketitle

\begin{abstract}
Diffusion models have made high-fidelity visual generation widely
accessible, while also lowering the barrier to producing convincing visual
forgeries and complicating content provenance. Existing watermarking
methods typically use a fixed embedding strength throughout denoising or
treat all watermark dimensions equally. As a result, they may introduce
unnecessary distortion when some watermark dimensions are already
reliable, while failing to sufficiently reinforce those that are more
vulnerable to subsequent perturbations. Therefore, robust watermarking
requires adapting both the embedding strength during denoising and the
optimization priority across watermark dimensions. To address this problem, we present SAMP, a self-adaptive framework
consisting of global margin regulation and vulnerability-sensitive
allocation. Global margin regulation adjusts the embedding strength
according to the separation between the target codeword and competing
codewords, together with the estimated effects of common attacks.
Vulnerability-sensitive allocation identifies watermark dimensions that
are more easily weakened by local perturbations and gives them higher
optimization priority. Experimental results demonstrate that SAMP achieves an average information recovery accuracy of 0.993. Compared with state-of-the-art methods, SAMP improves PSNR by 16.8\% and reduces LPIPS by 47.1\%. These results validate the effectiveness of
SAMP for reliable and high-quality diffusion watermarking.

\end{abstract}


\section{Introduction}
\label{sec:intro}
Diffusion models enable high quality visual content generation from
simple user inputs, while also lowering the barrier to producing
convincing visual forgeries. As generated content becomes increasingly
common, distinguishing its origin and verifying its authenticity become
more difficult. Digital watermarking provides a direct means of
associating provenance information with visual content, supporting source
tracing and authenticity verification.

Early diffusion watermarking methods embed watermark information by modifying the latent state that initializes the generation process. Tree Ring~\cite{TreeRing} introduces structured Fourier patterns into the initial noise for watermark verification. Gaussian Shading~\cite{GaussianShading}
encodes watermark information into latent representations while preserving the original latent distribution. RingID~\cite{RingID} extends latent-space
watermarking to multi-channel identification, while WIND~\cite{WIND} and PRC~\cite{PRC} explore latent-space embedding strategies for improving robustness and generation quality.

Recent methods further move watermark embedding into the denoising trajectory to achieve finer control over the generation process. LFLDW~\cite{LFLDW} injects watermark signals into selected latent-frequency components during intermediate denoising steps. GROW~\cite{GROW} extends this idea by progressively guiding the predicted clean latent across multiple denoising steps. However, existing trajectory-based methods still rely on predefined embedding strategies and cannot adapt the intervention strength to the
current watermark reliability or prioritize more vulnerable watermark dimensions. Therefore, robust diffusion watermarking requires adaptive control over both embedding strength and optimization priority.

In response to these challenges, we propose SAMP, a diffusion
watermarking framework that dynamically controls watermark embedding during the denoising process. Specifically, SAMP adjusts the watermark update process by controlling
the modification magnitude and the optimization priority of different watermark dimensions. SAMP's global margin regulation mechanism estimates the required latent modification by measuring the separation between the target watermark code and competing codewords. It further incorporates lightweight attack-aligned perturbations to improve the reliability of this estimation. The proposed vulnerability-sensitive allocation mechanism analyzes the sensitivity of different watermark dimensions to local perturbations and assigns higher optimization priority to vulnerable dimensions. Together, these designs reduce excessive latent perturbation while enhancing watermark robustness throughout generation.

% In response to these challenges, we propose an SAMP method. SAMP combines global margin regulation
% with vulnerability-sensitive allocation to adapt watermark embedding
% during diffusion denoising. Global margin regulation compares the target
% watermark code with competing codes and lightweight attack-aligned
% perturbations to determine how much the latent should be modified at the
% current denoising step. Vulnerability-sensitive allocation examines how
% different watermark components respond to local perturbations and gives
% greater optimization priority to those that are more easily degraded. The
% resulting latent update is applied before the next denoising step, reducing
% unnecessary modification while strengthening watermark information that
% remains vulnerable during generation.

The main contributions are summarized as follows:
\begin{itemize}
    \item A self-adaptive diffusion watermarking framework, SAMP, jointly regulates embedding strength and optimization priority during denoising.

    \item A global margin regulation mechanism that adjusts latent intervention according to codeword separation, attack-aligned perturbations, and modification cost.

    \item A vulnerability-sensitive allocation strategy that estimates coded-bit sensitivity through watermark-basis perturbations and adaptively reweights watermark constraints.
\end{itemize}

\begin{figure*}[t]
    \centering
    \includegraphics[
        width=0.95\textwidth,
        keepaspectratio,           % 保持宽高比
        trim=0 0 0 0,              % 如需裁剪白边：左 下 右 
        clip                       % 配合 trim 使用
    ]{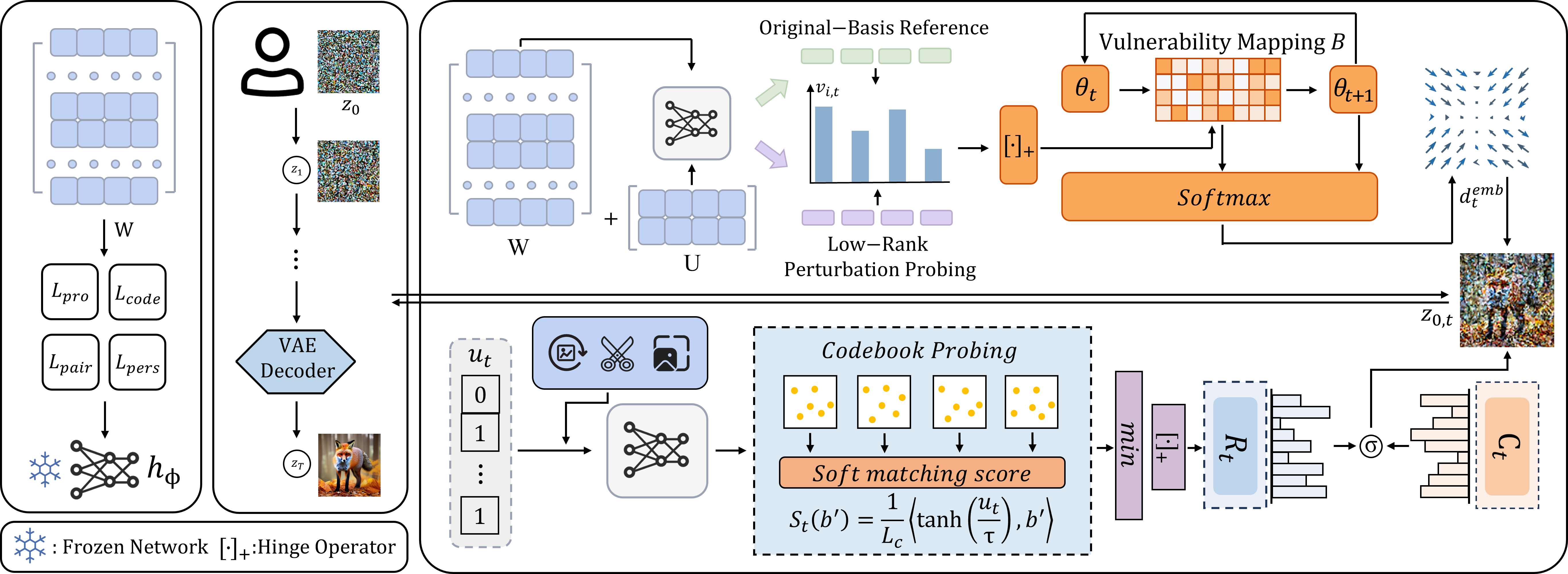}
    \caption{The framework of SAMP.}
    \label{fig:framework}
    \vspace{-2em} % 可选：减少底部空白，谨慎使用
\end{figure*}

\section{Proposed Method}
\label{sec:method}

As shown in Fig.~\ref{fig:framework}, SAMP is a diffusion watermarking method that combines global margin regulation and vulnerability sensitive allocation to improve watermark robustness while limiting unnecessary latent modification. Specifically, during generation, the predicted clean latent is mapped to a learned watermark coordinate space. The current embedding margin determines the watermark embedding strength, while local perturbations measure the local sensitivity of individual coded bits and adjust their optimization weights. The resulting magnitude and direction update the clean latent, which is then returned to the diffusion sampler.

\subsection{Watermark Representation}
\label{ssec:representation}

Let $\mathbf{m}\in\{0,1\}^{L}$ denote the length-$L$ secret information. An error-correction encoder $\operatorname{Enc}(\cdot)$ maps it to a bipolar coded watermark $\mathbf{b}=\operatorname{Enc}(\mathbf{m})\in\{-1,+1\}^{L_c}$, where $L_c>L$ is the coded length. Each $b_i$ specifies the target sign of one watermark coordinate. A key-dependent watermark basis $\mathbf{W}=[\mathbf{w}_1,\ldots,\mathbf{w}_{L_c}]\in\mathbb{R}^{d\times L_c}$ defines the latent-space writing directions, with $d$ the latent dimension. Its columns are orthonormalized, $\mathbf{W}^{\top}\mathbf{W}=\mathbf{I}$, to reduce inter-bit interference, and are then fixed for both training and generation.

Training samples are constructed offline by perturbing each unmarked latent along the watermark direction specified by $\mathbf{W}$ and $\mathbf{b}$, with the perturbation scaled by the latent norm and an imprint factor $\alpha$. The resulting marked and unmarked latents are used to train a nonlinear coordinate mapping $h_{\phi}: \mathbb{R}^{d} \rightarrow \mathbb{R}^{L_c}$. Given an input latent, $h_{\phi}$ produces $\mathbf{u} = h_{\phi}(\mathbf{z})$, and the embedding margin of the $i$-th coded bit is defined as $q_i(\mathbf{z}) = b_i u_i$. A positive $q_i$ indicates the target sign, while a larger $q_i$ indicates a larger margin against sign reversal. Thus, the coded watermark is represented by the sign constraints $b_i h_{\phi,i}(\mathbf{z}) > 0$ for $i = 1, \ldots, L_c$.

The coordinate mapping $h_{\phi}$ is learned offline by optimizing the composite objective:
\begin{equation}
\mathcal{L}_{\mathrm{off}} = \mathcal{L}_{\mathrm{code}} + \lambda_{\mathrm{pair}}\mathcal{L}_{\mathrm{pair}} + \lambda_{\mathrm{c}}\mathcal{L}_{\mathrm{probe}} + \lambda_{\mathrm{pers}}\mathcal{L}_{\mathrm{pers}} .
\label{eq:offline}
\end{equation}
Specifically, $\mathcal{L}_{\mathrm{code}}$ enforces correct codeword recovery from the marked latent $\mathbf{z}_w$; $\mathcal{L}_{\mathrm{pair}}$ penalizes cross-carrier coordinate discrepancies for the same watermark; $\mathcal{L}_{\mathrm{probe}}$ suppresses carrier-related information via an auxiliary probe $a(\cdot)$; and $\mathcal{L}_{\mathrm{pers}}$ improves coordinate stability under latent perturbations and spatial filtering.

Cropping and scaling are included among the perturbations used by $\mathcal{L}_{\mathrm{pers}}$ to ensure coordinate stability under spatial changes; after training, $h_{\phi}$ is frozen for generation and extraction.

\subsection{Global Margin Regulation}
\label{ssec:margin_regulation}

Global margin regulation sets the embedding strength per denoising step from the current margin, ensuring sufficient decoding margin without excessive latent modification. Rather than independent sign-constraint evaluation, SAMP measures target-to-competitor separation: a large global margin indicates reliability, whereas a small one requires intervention.

At the $t$-th denoising step, the diffusion model predicts the clean latent from the current noisy state:
\begin{equation}
\hat{\mathbf{z}}_{0,t} = \frac{\mathbf{z}_t - \sqrt{1-\bar{\alpha}_t}\,\boldsymbol{\epsilon}_{\theta}(\mathbf{z}_t,t)}{\sqrt{\bar{\alpha}_t}}.
\label{eq:z0}
\end{equation}
The predicted clean latent is mapped to the fixed watermark coordinate space as $\mathbf{u}_t=h_{\phi}(\hat{\mathbf{z}}_{0,t})$. Let $\mathcal{C}$ denote the set of valid bipolar codewords generated by the encoder. For a candidate codeword $\mathbf{b}'\in\mathcal{C}$, we measure its similarity to the current coordinates by
\begin{equation}
S_t(\mathbf{b}') = \frac{1}{L_c} \left\langle \tanh\!\left(\frac{\mathbf{u}_t}{\tau}\right), \mathbf{b}' \right\rangle,
\label{eq:score}
\end{equation}
where $\tau$ is a temperature parameter. The separation between the target codeword and its strongest competitor is
\begin{equation}
G_{\mathrm{code},t} = S_t(\mathbf{b}) - \max_{\substack{\mathbf{b}'\in\mathcal{C}\\\mathbf{b}'\neq\mathbf{b}}} S_t(\mathbf{b}').
\label{eq:code_gap}
\end{equation}

Since the latent distribution evolves during denoising and post-generation, the codeword margin alone may overestimate the actual robustness. SAMP therefore evaluates the margin over lightweight latent proxies with predefined transformations to approximate potential perturbations. Let $G_{\mathrm{proxy},t}$ denote the minimum target-to-competitor separation among these proxies. The conservative margin and the corresponding robustness gap are defined as
\begin{equation}
G_t = \min \left\{ G_{\mathrm{code},t}, G_{\mathrm{proxy},t} \right\}, \qquad
R_t = \left[g_{\mathrm{safe}} - G_t\right]_+,
\label{eq:global_margin}
\end{equation}
where $g_{\mathrm{safe}}$ is the desired safety margin. A larger $R_t$ indicates a larger deficiency from the target robustness level and thus requires stronger watermark intervention.

However, $R_t$ only measures the required embedding strength and ignores the modification difficulty of the current latent state. After determining the writing direction, SAMP estimates the local modification cost by applying a small directional displacement followed by the same radius-preserving projection as the final update. The resulting relative latent variation is used as the cost $C_t$, which is normalized by a calibration scale to obtain $\bar{C}_t$.

The step-wise intervention magnitude is then determined by
\begin{equation}
\eta_t = \eta_t^{\mathrm{eff}} \sigma\!\left( \lambda_Q R_t-\bar C_t \right),
\label{eq:strength}
\end{equation}
Here, $\sigma(\cdot)$ is the sigmoid function, $\lambda_Q$ controls the response to the remaining robustness requirement, and $\eta_t^{\mathrm{eff}}$ is the scheduled intervention scale over the active writing interval. Thus, SAMP strengthens intervention when the global margin is insufficient and suppresses unnecessary modification when modification is costly.

\begin{table*}[t]
\caption{
Comparison of different watermarking methods on two datasets.
The upper and lower parts correspond to Stable-Diffusion-Prompts and MS-COCO datasets, respectively.
For imperceptibility metrics (PSNR, SSIM, FID, LPIPS), we report the performance on clean images.
For robustness, we report the M-ACC (\%) under various attacks.
Higher values indicate better performance for metrics with $\uparrow$, while lower values indicate better performance for metrics with $\downarrow$.
The best and second-best results are highlighted in bold and underlined, respectively.
}
\label{tab:performance}

\centering
\small
\setlength{\tabcolsep}{4pt}
\renewcommand{\arraystretch}{0.95}

\begin{tabular}{l cccc ccccccccc}

\toprule

\textbf{Method} &
\multicolumn{4}{c}{\textbf{Imperceptibility}} &
\multicolumn{9}{c}{\textbf{Robustness (M-ACC $\uparrow$)}}\\

\cmidrule(lr){2-5}
\cmidrule(lr){6-14}

& PSNR$\uparrow$
& SSIM$\uparrow$
& LPIPS$\downarrow$
& FID$\downarrow$
& Clean
& JPEG
& Noise
& Blur
& Crop
& Rotation
& Bright
& Adversarial
& Avg.

\\

\midrule

Tree-Ring
&15.96&0.537&0.473&25.69
&0.98&0.87&0.84&0.27&0.06&0.03&0.84&0.72&0.576\\

Gaussian Shading
&23.68&0.811&0.093&24.62
&1.00&1.00&1.00&0.91&0.92&0.43&0.98&0.82&0.883\\

RingID
&13.20&0.453&0.533&25.95
&1.00&1.00&\underline{1.00}&
\textbf{1.00}&0.23&\textbf{1.00}&0.99&0.81&0.879\\

WIND
&14.71&0.541&0.471&22.06
&1.00&1.00&0.97&0.95&0.77&0.76&0.99&0.88&0.915\\

GROW
&\underline{27.37}&\underline{0.824}&\underline{0.087}&\underline{14.59}
&\underline{1.00}&\underline{1.00}&0.89&0.92&
\underline{1.00}&0.97&\underline{1.00}&
\underline{0.92}&\underline{0.963}\\

\textbf{SAMP}
&\textbf{31.98}&\textbf{0.901}&\textbf{0.046}&\textbf{9.52}
&\textbf{1.00}&\textbf{1.00}&\textbf{1.00}&
\underline{0.99}&\textbf{1.00}&
\underline{0.98}&\textbf{1.00}&
\textbf{0.97}&\textbf{0.993}\\

\midrule

Tree-Ring
&16.52&0.521&0.433&26.14
&0.93&0.63&0.72&0.49&0.05&0.04&0.68&0.39&0.491\\

Gaussian Shading
&23.96&0.799&0.094&25.34
&1.00&1.00&\underline{0.98}&0.90&
0.94&0.57&0.99&0.93&0.914\\

RingID
&14.19&0.443&0.426&26.41
&1.00&1.00&0.93&
\underline{1.00}&0.54&
\underline{1.00}&0.95&0.85&0.909\\

WIND
&14.33&0.512&0.411&23.57
&1.00&1.00&0.95&0.99&
0.84&0.76&\underline{1.00}&0.88&0.928\\

GROW
&\underline{27.67}&\underline{0.839}&\underline{0.061}&
\underline{15.34}
&\underline{1.00}&\underline{1.00}&0.92&0.99&
\textbf{1.00}&\textbf{1.00}&0.99&
\underline{0.94}&\underline{0.980}\\

\textbf{SAMP}
&\textbf{32.18}&\textbf{0.907}&\textbf{0.041}&
\textbf{8.25}
&\textbf{1.00}&\textbf{1.00}&\textbf{0.99}&
\textbf{1.00}&\underline{0.98}&0.96&
\textbf{1.00}&\textbf{0.99}&
\textbf{0.990}\\

\bottomrule

\end{tabular}
\end{table*}

\subsection{Vulnerability Sensitive Allocation}
\label{ssec:vulnerability}

The required modification varies across watermark constraints, as different sign constraints exhibit different sensitivities to local latent variations. SAMP therefore estimates constraint-level vulnerability to adapt the optimization priority.

Specifically, SAMP perturbs the watermark basis as $\mathbf{W}_k=\mathbf{W}+\mathbf{U}\mathbf{A}_k$, where $\mathbf{U}\in\mathbb{R}^{d\times r}$ spans structured directions orthogonal to $\mathbf{W}$ ($\mathbf{W}^{\top}\mathbf{U}=\mathbf{0}$), and $\mathbf{A}_k\in\mathbb{R}^{r\times L_c}$ denotes the coefficients of the $k$-th perturbation. The resulting $\mathbf{W}_k$ provides local variations around the original writing basis for evaluating the sensitivity of individual coded bits. The corresponding latent state and vulnerability are computed as

\begin{equation}
\hat{\mathbf{z}}_{0,t}^{(k)} = \hat{\mathbf{z}}_{0,t} - \delta_t \frac{\mathbf{W}_k \mathbf{b}}{\| \mathbf{W}_k \mathbf{b} \|_2 + \epsilon}, \qquad
v_{i,t} = \frac{1}{K} \sum_{k=1}^{K} \left[ q_{i,t} - q_{i,t}^{(k)} \right]_+,
\label{eq:vulnerability}
\end{equation}
where $\delta_t$ controls the perturbation magnitude. $h_{\phi}$ maps each perturbed latent $\hat{\mathbf{z}}_{0,t}^{(k)}$ to the embedding margin $q_{i,t}^{(k)}$ of coded bit $i$. Averaging over $K$ perturbations gives $v_{i,t}$; a larger value indicates higher sensitivity of bit $i$ to local latent changes.

Constraint optimization weights $\mathbf{w}_t$ are updated online using orthogonal subspace matrix $\mathbf{B}$ and state variable $\boldsymbol{\theta}_t$:
\begin{equation}
\boldsymbol{\theta}_{t+1} = \boldsymbol{\theta}_t + \rho_{\theta} \mathbf{B}^{\top} (\mathbf{v}_t - \bar{v}_t \mathbf{1}), \qquad \mathbf{w}_t = \operatorname{Clip}\left( \operatorname{softmax}(\mathbf{B}\boldsymbol{\theta}_{t+1}) \right),
\label{eq:weights}
\end{equation}
where $\bar{v}_t$ is mean vulnerability. Evaluating the weighted loss $\mathcal{L}_t = \sum_i w_{i,t} \operatorname{softplus}(
\gamma_t - q_{i,t})$ yields the normalized writing direction
\begin{equation}
\mathbf{d}_t^{\mathrm{emb}} = -\frac{\nabla_{\hat{\mathbf{z}}_{0,t}}\mathcal{L}_t}{\|\nabla_{\hat{\mathbf{z}}_{0,t}}\mathcal{L}_t\|_2 + \epsilon}.
\label{eq:direction}
\end{equation}

The magnitude $\eta_t$ from Sec.~\ref{ssec:margin_regulation} is then applied along $\mathbf{d}_t^{\mathrm{emb}}$ to derive the latent update $\Delta\mathbf{z}_t = \eta_t \|\hat{\mathbf{z}}_{0,t}\|_2 \mathbf{d}_t^{\mathrm{emb}}$. Since this update can change both the direction and the norm of the predicted clean latent, SAMP preserves the original latent scale after the watermark modification. A radius-preserving projection $\Pi_{\mathrm{rad}}$ therefore produces the updated clean latent $\hat{\mathbf{z}}_{0,t}^{+} = \Pi_{\mathrm{rad}}(\hat{\mathbf{z}}_{0,t} + \Delta\mathbf{z}_t)$, which is returned to the diffusion sampler for the next denoising iteration.

\subsection{Watermark Decoding}
\label{ssec:decoding}

Given a received image $\mathbf{x}'$, SAMP recovers the secret message without inversion. The VAE-encoded latent is mapped by the frozen coordinate mapping $h_{\phi}$ to watermark coordinates $\mathbf{u}' = h_{\phi}(\operatorname{VAEEnc}(\mathbf{x}'))$. The nearest valid codeword in $\mathcal{C}$ is then found via the matching score $S(\cdot)$ in Eq.~\eqref{eq:score}:
\begin{equation}
\hat{\mathbf{b}} = \arg\max_{\mathbf{b}'\in\mathcal{C}} S(\mathbf{u}', \mathbf{b}'), \qquad \hat{\mathbf{m}} = \operatorname{Dec}(\hat{\mathbf{b}}).
\label{eq:decode}
\end{equation}
This extraction relies solely on one VAE encoding and the coordinate network.

\section{Experiments}
\label{sec:experiments}

\subsection{Experimental Setup}
\label{ssec:setup}

\textbf{Dataset.} We evaluate SAMP on two text-to-image datasets with different text distributions. Stable-Diffusion-Prompts~\cite{SDPrompts} contains diverse, open-ended user-generated prompts, while MS-COCO~\cite{MSCOCO} provides concise, natural human-written captions paired with real images. For each dataset, 5,000 text inputs are randomly selected and used without modification.

\textbf{Baselines.} We compare SAMP with five representative diffusion watermarking methods,
including Tree-Ring~\cite{TreeRing}, Gaussian Shading~\cite{GaussianShading},
RingID~\cite{RingID}, WIND~\cite{WIND}, and GROW~\cite{GROW}. All methods are evaluated under the same experimental settings.

\textbf{Implementation Details.}
All experiments are conducted on Stable Diffusion 2.1~\cite{SD21}. Images are generated at $512\times512$ resolution using 50 DDIM steps ($T=50$) and a guidance scale of 7.5. SAMP embeds $L=16$ bits into an $L_c=32$-symbol RM$(2,5)$ codeword during the final $T_e=25$ denoising steps. The coordinate mapping $h_\phi$ is pretrained offline using 2,000 cached latents with Adam for 5,000 steps ($\text{lr}=10^{-4}$, batch size=32), followed by crop-resize fine-tuning for 3,000 steps with learning rate $2\times10^{-5}$. During embedding, $\gamma_t$ is linearly scheduled from 0.10 to 1.00, with temperature $\tau=1$ and vulnerability probing interval of five steps ($K=2$). All methods use the same prompts and initial latents for fair comparison.

\textbf{Metrics.}
The performance of all methods is evaluated in terms of imperceptibility and robustness. Imperceptibility is measured by PSNR, SSIM, LPIPS~\cite{zhang2018perceptual},
and FID~\cite{heusel2017fid} on generated images. Robustness is measured by complete message accuracy (M-ACC), which denotes the ratio of images with successfully recovered messages under different attacks.

\subsection{Main Results}
\label{ssec:results}
\begin{figure}[t]
\centering
\includegraphics[
width=\columnwidth,
trim=70 0 0 0,
clip
]{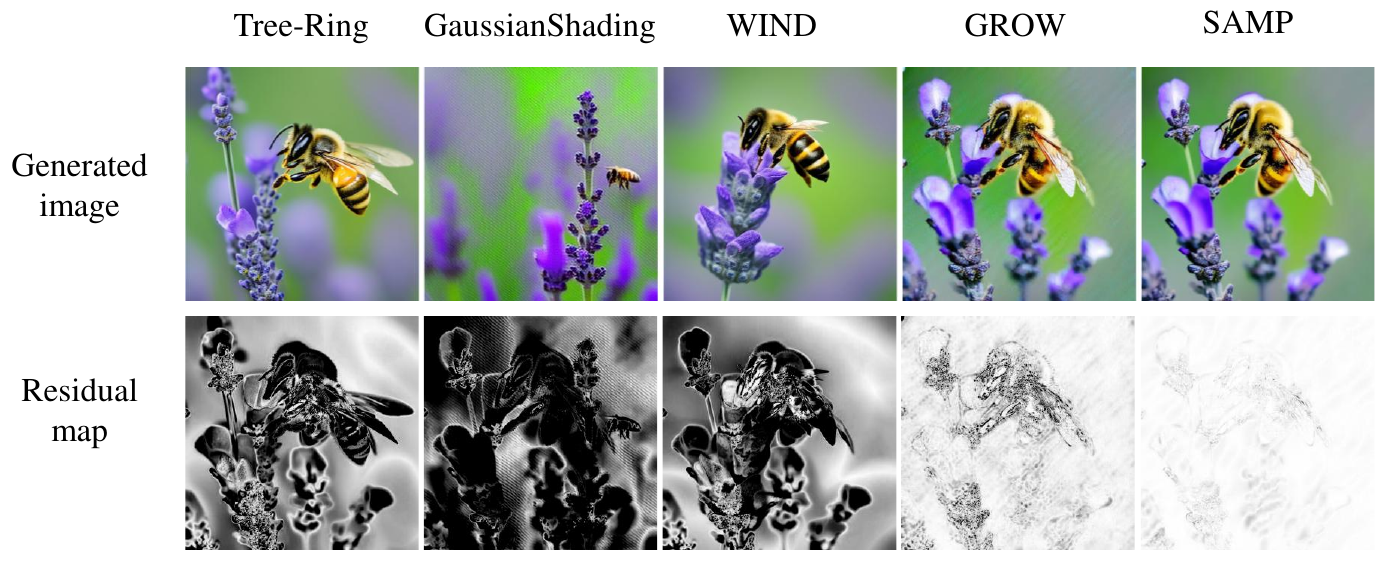}

\vspace{-4mm}

\caption{
Qualitative comparison of generated images and residual maps under the prompt
\textit{“A macro photo of honeybee hovering on lavender flower, sharp details.”}
The first row shows the generated images, while the second row presents the
corresponding residual maps for different watermarking methods.
}
\label{fig:photos}
\end{figure}

\textbf{Imperceptibility.} Table~\ref{tab:performance} shows SAMP's superior imperceptibility across datasets. On Stable-Diffusion-Prompts, SAMP achieves PSNR 31.98~dB, SSIM 0.901, LPIPS 0.046, and FID 9.52. Compared with GROW, it improves PSNR by 4.61~dB (16.8\%) and reduces LPIPS by 47.1\%, indicating substantially less distortion. This mainly stems from global margin regulation, which adaptively reduces latent intervention when watermark reliability is sufficient. Fig.~\ref{fig:photos} qualitatively compares generated images; SAMP maintains visual consistency and produces no noticeable watermark artifacts.

\textbf{Robustness.} Table~\ref{tab:performance} demonstrates the robustness of SAMP
against diverse attacks, including image transformations and diffusion-based regeneration. SAMP achieves an average M-ACC of 0.993 across all eight evaluation conditions, maintaining near-perfect information recovery under common perturbations including JPEG compression, Gaussian noise, cropping, and brightness adjustment. For more challenging attacks, SAMP maintains M-ACC values of 0.98 and 0.97 on rotation and regeneration, respectively, outperforming the strongest baseline by 5.4\% under regeneration. Such robustness benefits from the vulnerability-sensitive allocation mechanism, which identifies fragile watermark constraints and assigns higher optimization priority to preserve watermark reliability against local perturbations.

\subsection{Computational Cost} Table~\ref{tab:time} reports the generation and extraction latency of different watermarking methods. SAMP requires 4.35~s for image generation, with only 0.21~s introduced by watermark embedding, while achieving an extraction time of 0.050~s per image. Unlike inversion-based extraction pipelines, SAMP directly decodes the watermark through a single VAE encoding and the frozen coordinate mapping $h_\phi$, avoiding iterative diffusion inversion and substantially reducing extraction overhead.

\begin{table}[t]
\caption{Runtime evaluation per image. Generation shows total time / watermarking overhead.}
\label{tab:time}
\centering
\small
\setlength{\tabcolsep}{6pt}
\begin{tabular}{lcc}
\hline
\textbf{Method} & \textbf{Generation Time (s)} & \textbf{Extraction Time (s)} \\
\hline
Tree-Ring        & \textbf{4.15 / 0.01} & 2.413 \\
Gaussian Shading & 4.37 / 0.23          & 2.404 \\
RingID           & \underline{4.17 / 0.03}          & 2.502 \\
WIND             & 4.30 / 0.16          & 2.453 \\
GROW             & 4.28 / 0.14          & \underline{0.079} \\
\textbf{SAMP}    & 4.35 / 0.21          & \textbf{0.050} \\
\hline
\end{tabular}
\end{table}

\subsection{Ablation Study}
\label{ssec:ablation}
Table~\ref{tab:ablation} evaluates the two SAMP components. The variant with neither uses fixed embedding, achieving PSNR 23.40~dB and M-ACC 0.962. Global margin regulation alone raises PSNR to 32.24~dB by reducing unnecessary latent intervention when the watermark margin is sufficient, but lowers M-ACC to 0.911 because vulnerable constraints are not explicitly reinforced. Vulnerability-sensitive allocation alone improves M-ACC to 0.984 by prioritizing fragile constraints. Combining both achieves PSNR 31.98~dB and M-ACC 0.993, showing that they jointly improve imperceptibility and robustness.

\begin{table}[t]
\caption{Ablation of margin regulation and vulnerability allocation.}
\label{tab:ablation}
\centering
\small
\setlength{\tabcolsep}{6pt}
\begin{tabular}{lcc}
\hline
\textbf{Variant} & PSNR$\uparrow$ & M-ACC$\uparrow$ \\
\hline
Fixed embedding & 23.40 & 0.962 \\
Global margin regulation only & \textbf{32.24} & 0.911 \\
Vulnerability allocation only & 23.32 & \underline{0.984} \\
\textbf{SAMP} & \underline{31.98} & \textbf{0.993} \\
\hline
\end{tabular}
\end{table}

\section{Conclusion}
\label{sec:conclusion}
We presented SAMP, an adaptive diffusion watermarking framework that regulates latent intervention during denoising. By combining message-level margin metrics with counterfactual sensitivity probes, SAMP dynamically modulates embedding intensity and constraint allocation. Experimental results show that SAMP reduces LPIPS by 47.1\% over the strongest baseline and achieves an average bit recovery accuracy of 0.993 across diverse image degradations with single-pass VAE decoding.

\bibliographystyle{IEEEbib}
\bibliography{strings,refs}

\end{document}